\documentclass[10pt,twocolumn,letterpaper]{article}

\usepackage{cvpr}
\usepackage{times}
\usepackage{epsfig}
\usepackage{graphicx}
\usepackage{amsmath}
\usepackage{amssymb}

\usepackage{tabularx}
\usepackage{booktabs}
\usepackage{multirow}

\usepackage[pagebackref=true,breaklinks=true,letterpaper=true,colorlinks,bookmarks=false]{hyperref}

\cvprfinalcopy 

\def\cvprPaperID{244} 

\newcommand{\argmax}{\operatornamewithlimits{arg\,max}}

\ifcvprfinal\fi
\begin{document}

\title{Weakly Supervised Action Learning with RNN based Fine-to-coarse Modeling}

\author{Alexander Richard, Hilde Kuehne, Juergen Gall\\
University of Bonn, Germany\\
{\tt\small \{richard,kuehne,gall\}@iai.uni-bonn.de}
}

\maketitle

\begin{abstract}

We present an approach for weakly supervised learning of human actions. Given a set of videos and an ordered list of the occurring actions, the goal is to infer start and end frames of the related action classes within the video and to train the respective action classifiers without any need for hand labeled frame boundaries. 
To address this task, we propose a combination of a discriminative representation of subactions, modeled by a recurrent neural network, and a coarse probabilistic model to allow for a temporal alignment and inference over long sequences. While this system alone already generates good results, we show that the performance can be further improved by approximating the number of subactions to the characteristics of the different action classes. To this end, we adapt the number of subaction classes by iterating realignment and reestimation during training.
The proposed system is evaluated on two benchmark datasets, the Breakfast and the Hollywood extended dataset, showing a competitive performance on various weak learning tasks such as temporal action segmentation and action alignment.
    
\end{abstract}

%


\section{Introduction}

Given the large amount of available video data, \eg on Youtube, from movies or even in the context of surveillance, methods to automatically find and
classify human actions within these videos gained an increased interest within the
last years~\cite{wang2013action, karpathy2014large, simonyan2014two, richard2016temporal, yeung2016endtoend}.

While there are several successful methods to classify trimmed video clips~\cite{wang2013action, simonyan2014two},
temporal localization and classification of human actions in untrimmed, long video sequences are
still a huge challenge. Most existing approaches in this field rely on fully annotated
video data, \ie the exact start and end time of each action in the training set needs
to be provided~\cite{rohrbach2012database, richard2016temporal, yeung2016endtoend}.
For real world applications, this requires an enormous effort of creating training data and can be too expensive to realize.
\begin{figure}[t]
    \centering
    \includegraphics[width=0.45\textwidth]{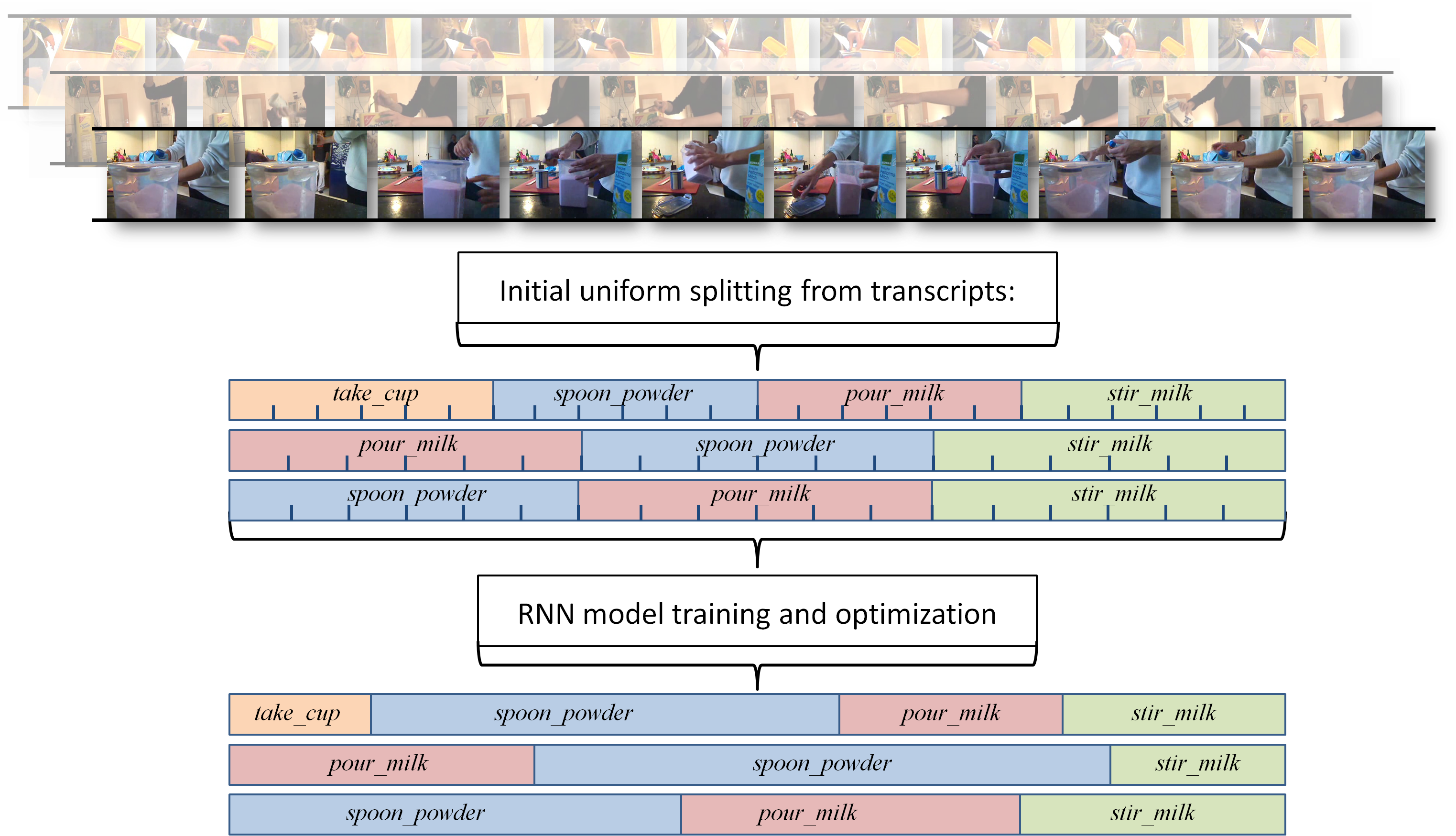}
    \caption{Overview of the proposed weak learning system. Given a list of ordered actions for each video, an initial segmentation is generated
by uniform segmentation. Based on this input information we iteratively train an RNN-based fine-to-coarse system to align the frames to the respective action.}
    \label{fig:system_overview}
\vspace{-3mm}
\end{figure}
Therefore, weakly supervised methods are of particular interest. Such methods usually
assume that only an ordered list of actions occurring in the video is annotated instead
of exact framewise start and end points~\cite{duchenne09automatic, bojanowski14weakly, huang2016connectionist}.
This information is much easier to generate for human annotators, or can even be automatically derived from scripts~\cite{laptev08learning,marszalek09actions} or subtitles~\cite{alayrac16unsupervised}.
The idea that all those approaches share is that - given a set of videos and a respective list of the actions that occur in the video - it is possible to learn the characteristics of the related action classes, to infer their start and end frames within the video, and to build the corresponding action models without any need for hand labeled frame boundaries (see Figure \ref{fig:system_overview}).

In this work, we address the task of weak learning of human actions by a fine-to-coarse model. On the fine grained level, we use a discriminative representation of subactions, modeled by a recurrent neural network as \eg used by \cite{donahue15longterm, ng15beyond, singh16multistream, wu15modeling}. In our case, the RNN is used as basic recognition model as it provides robust classification of small temporal chunks. This allows to capture local temporal information. The RNN is supplemented by a coarse probabilistic model to allow for temporal alignment and inference over long sequences.

Further, to bypass the difficulty of modeling long and complex action classes, we divide all actions into smaller building blocks.
Those subactions are eventually modeled within the RNN and later combined by the inference process. The usage of subactions allows
to distribute heterogeneous information of one action class over many subclasses and to capture characteristics such as the length of the overall action class. Additionally, we show that automatically learning the number of subactions for each action class leads to a notably improved performance.

Our model is trained with an iterative procedure. Given the weakly supervised training data, an initial segmentation is generated
by uniformly distributing all actions among the video. For each obtained action segment, all subactions are then also uniformly
distributed among the part of the video belonging to the corresponding action. This way, an initial alignment between video frames
and subactions is defined. In an iterative phase, the RNN is then trained on this alignment and used in combination with the
coarse model to infer new action segment boundaries. From those boundaries, we recompute the number of subactions needed for each action class, distribute them again among the frames aligned to the respective action, and repeat the training process until convergence.

We evaluate our approach on two common benchmark datasets, the Breakfast dataset \cite{kuehne14language} and the Hollywood extended dataset \cite{bojanowski14weakly}, regarding two different tasks. The first task is temporal action segmentation, which refers to a combined segmentation and classification, where the test video is given without any further annotation. The second task is aligning a test video to a given order of actions, as proposed by Bojanowski \etal  \cite{bojanowski14weakly}. Our approach is able to outperform current state-of-the-art methods on both tasks.


\section{Related Work}

For the case of fully supervised learning of actions, well-studied deep learning and
temporal modeling approaches exist.
While the authors of ~\cite{yeung2016endtoend} focus on a purely neural network based approach,
Tang \etal~\cite{tang2012learning} propose to learn the latent temporal structure of videos
with a hidden Markov model. Combining deep learning and temporal modeling, the authors
of~\cite{lea2016segmental} use a segmental CNN and a semi-Markov model to represent temporal
transitions between actions. However, these methods are not applicable in a weakly supervised
setting.

Addressing the problem of weakly supervised learning of actions, a variety of different approaches have been explored.
First works, proposed by Laptev \etal~\cite{laptev08learning} and Marszalek \etal~\cite{marszalek09actions}, focus on mining training samples from movie scripts. They extract class samples based on the respective text passages and use those snippets for training without applying a dedicated temporal alignment of the action within the extracted clips. 
First attempts for learning action classes including temporal alignment on weakly annotated data are made by Duchenne \etal \cite{duchenne09automatic}. 
Here, it is assumed that all snippets contain only one class and the task is to temporally segment frames containing the relevant action from the background activities. The temporal alignment is thus interpreted as a binary clustering problem, separating temporal snippets containing the action class from the background segments. The clustering problem is formulated as a minimization of a discriminative cost function.
This problem formulation is extended by Bojanowski \etal~\cite{bojanowski14weakly} also introducing the Hollywood extended dataset. Here, the weak learning is formulated as a temporal assignment problem. Given a set of videos and the action order of each video, the task is to assign the respective class to each frame, thus to infer the respective action boundaries. The authors propose a discriminative clustering model using the temporal ordering constraints to combine classification of each action and their temporal localization in each video clip. They propose the usage of the Frank-Wolfe algorithm to solve the convex minimization problem. This method has been adopted by Alayrac \etal \cite{alayrac16unsupervised} for unsupervised learning of task and story lines from instructional video.
Another approach for weakly supervised learning from temporally ordered action lists is introduced by Huang \etal~\cite{huang2016connectionist}. They feature extended connectionist temporal classification and propose the induction of visual similarity measures to prevent the CTC framework from degeneration and to enforce visually consistent paths. 
On the other hand, Kuehne \etal~\cite{kuehne2016weakly} borrow on the concept of flat models in speech recognition. They model actions by hidden Markov models (HMMs) and aim to maximize the probability of training sequences being generated by the HMMs by iteratively inferring the segmentation boundaries for each video and using the new segmentation to reestimate the model. The last two approaches were both evaluated on the Hollywood extended as well as on the Breakfast dataset, thus, these two datasets are also used for the evaluation of the here proposed framework.

Beside the approaches focusing on weak learning of human actions based on temporally ordered labels, also other weak learning scenarios have been explored.
A closely related approach comes from the field of sign language recognition. Here, Koller \etal~\cite{koller16deephand} integrate CNNs with hidden Markov models to learn sign language hand shapes based on a single frame CNN model from weakly annotated data. They evaluate their approach on various large scale sign language corpora, \eg for Danish and New Zealand sign language.
Gan \etal~\cite{gan2016webly} show an approach to learn action classes from web images and videos retrieved by specific search queries. They feature a pairwise match of images and video frames and combine this with a regularization over the selected video frames to balance the matching procedure. The approach is evaluated on standard action classification datasets such a UCF101 and Trecvid.
Also learning from web videos and images is the approach of \cite{sun15temporal}. Weak video labels and noisy image labels are taken as input, and localized action frames are generated as output. The localized action frames are used to train action recognition models with long short-term memory networks. Results are reported, among others, for temporal detection on the THUMOS 2014 dataset.
Another idea is proposed by Misra \etal~\cite{misra2016shuffle}, aiming to learn a temporal order verification for human actions in an unsupervised way by training a CNN with correct vs.\ shuffled video snippets and thus capturing temporal information. The system can be used for pre-training feature extractors on small datasets as well as in combination with other supervised methods.
A more speech related task is also proposed by Malmaud \etal~\cite{malmaud15what}, trying to align recipe steps to automatically generated speech transcripts from cooking videos. They use an hybrid HMM model in combination with a CNN based visual food detector to align a sequence of instructions, e.g. from textual recipes, to a video of someone carrying out a task.
Finally,~\cite{wu15watch} propose an unsupervised technique to derive action classes from RGB-D videos, respectively human skeleton representations, also considering an activity as a sequence of short-term action clips. They propose Gibbs sampling for learning and inference of long activities from basic action words and evaluate their approach on an RGB-D activity video dataset.


\section{Technical Details}

In the following, we describe the proposed framework in detail, starting with a short definition of the weak learning task and the related training data. We then define our model and describe the overall training procedure as well as how it can be used for inference.

\subsection{Weakly Supervised Learning from Action Sequences}

In contrast to fully supervised action detection or segmentation approaches, where frame based ground truth data is available, in weakly supervised learning only an ordered list of the
actions occurring in the video is provided for training. 
A video of making tea, for instance, might consist of taking a cup, putting the teabag in it, and pouring water into the cup.
While fully supervised tasks would provide a temporal annotation of each action start and end time, in our weakly supervised setup, all given information is the ordered
action sequence
\begin{align*}
    \text{\texttt{take\_cup, add\_teabag, pour\_water}}.
\end{align*}

More formally, we assume the training data is a set of tupels $ (\mathbf{x}_1^T, \mathbf{a}_1^N ) $,
where $ \mathbf{x}_1^T $ are framewise features of a video with $ T $ frames and
$ \mathbf{a}_1^N $ is an ordered sequence $ (a_1, \dots, a_N) $ of actions occurring
in the video. The segmentation of the video is defined by the mapping
\begin{align}
    n(t): \{1,\dots,T\} \mapsto \{1,\dots,N\}
    \label{mapping}
\end{align}
that assigns an action segment index to each frame. Since our model iteratively optimizes
the action segmentation, initially, this can simply be a linear segmentation of the
provided actions, see Figure~\ref{fig:train}a.
The likelihood of the video frames $ \mathbf{x}_1^T $ given the action transcripts
$ \mathbf{a}_1^N $ is then defined as
\begin{align}
    p(\mathbf{x}_1^T | \mathbf{a}_1^N) := \prod_{t=1}^T p\big(x_t|a_{n(t)}\big),
\end{align}
where $ p(x_t|a_{n(t)}) $ is the probability of frame $ x_t $ being generated by the action $ a_{n(t)} $.

The action classes given for training usually describe longer, task-oriented procedures that naturally consist of more than one significant motion, e.g. \textit{take\_cup} can involve moving a hand towards a cupboard, opening the cupboard, grabbing the cup and placing it on the countertop. This makes it difficult to train long, heterogeneous actions as a whole. To efficiently capture those characteristics, we propose to model each action as a sequential combination of subactions.
Therefore, for each action class $ a $, a set of subactions $ s_1^{(a)},\dots,s_{K_a}^{(a)} $ is defined. The number $ K_a $ is
initially estimated by a heuristic and refined during the optimization process.
Practically, this means that we subdivide the original long action classes into a set of smaller subactions. As subactions are obviously not defined by the given ordered action sequences, we treat them as latent variables that need to be learned by the model. In the following system description, we assume that the subaction frame boundaries are known, \eg from previous iterations or from an initial uniform segmentation (see Figure~\ref{fig:train}b), and discuss the inference of more accurate boundaries in Section \ref{sec:inference}.

\subsection{Coarse Action Model}

In order to combine the fine grained subactions to action sequences, a hidden Markov model
$ \mathcal{H}_a $ for each action $ a $ is defined. The HMM ensures that subactions
only occur in the correct ordering, \ie that $ s_i^{(a)} \prec s_j^{(a)} $ for $ i \leq j $.
More precisely, let
\begin{align}
    s(t): \{1, \dots, T\} \mapsto \{s_1^{(a_1)},\dots,s_{K_{a_N}}^{(a_N)}\}
\end{align}
be the known mapping from video frames to the subactions of the ordered action sequence
$ \mathbf{a}_1^N $. This is basically the same mapping as the one in Equation~\eqref{mapping}
but on subaction level rather than on action level.
When going from one frame to the next, we only allow to assign either the same subaction or
the next subaction, so if at frame $ t $, the assigned subaction is $ s(t) = s^{(a)}_i $,
then at frame $ t + 1 $, either $ s(t+1) = s^{(a)}_i $ or $ s(t+1) = s^{(a)}_{i+1} $.
The likelihood of the video frames $ \mathbf{x}_1^T $ given the action transcripts
$ \mathbf{a}_1^N $ is then
\begin{align}
    p(\mathbf{x}_1^T | \mathbf{a}_1^N) :=
        \prod_{t=1}^T p\big(x_t|s(t)\big) \cdot p\big(s(t)|s(t-1)\big),
        \label{coarseModel}
\end{align}
where $ p(x_t|s) $ are probabilities computed by the fine-grained model, see Section~\ref{sec:finegrained}.
The transition probabilities $ p(s|s') $ from subaction $ s' $ to subaction $ s $
are relative frequencies of how often the transition $ s' \rightarrow s $ occurs
in the $ s(t) $-mappings of all training videos.




\subsection{Fine-grained Subaction Model}
\label{sec:finegrained}

\begin{figure}[tb]
    \centering
    \includegraphics{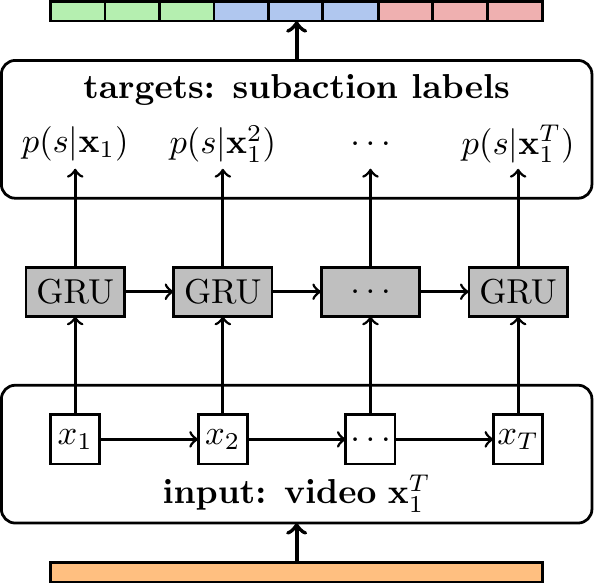}
    \caption{RNN using gated recurrent units with framewise video features as input.
             At each frame, the network outputs a probability for each possible subaction
             while considering the context of the video.}
    \label{fig:rnntrain}
    \vspace{-3mm}
\end{figure}

For the classification of fine-grained subactions, we use an RNN with a single hidden layer of gated recurrent units (GRUs) \cite{cho2014translation}. It is a simplified version of LSTMs that shows comparable performance~\cite{jozefowicz15empirical,chung2014empirical} also in case of video classification~\cite{ballas16delving}. The network is shown in Figure \ref{fig:rnntrain}. 

For each frame, it predicts a probability distribution over all subactions, while the recurrent structure of the network allows to incorporate local temporal context. Since the RNN generates a posterior distribution
$ p(s|x_t) $ but our coarse model deals with subaction-conditional probabilities,
we use Bayes' rule to transform the network output to
\begin{align}
    p(x_t|s) = \mathrm{const} \cdot \frac{p(s|x_t)}{p(s)}. \label{bayes}
\end{align}

\noindent
\textbf{Solving Efficiency Issues.}
Recurrent neural networks are usually trained using backpropagation through time
(BPTT) \cite{werbos1990backpropagation}, which requires to process the whole sequence in a forward
and backward pass. As videos can be very long and may easily exceed $ 10,000 $ frames,
the computation time per minibatch can be extremely high.
Even worse, long videos may not fit the memory of high-end GPUs, since during training the output of all network layers needs to be stored for each frame of the video in order to compute the gradient.

We tackle this problem by using small chunks around each video frame that can
be processed efficiently and with a reasonably large minibatch size in order to
enable efficient RNN training on long videos. For each
frame $ t $, we create a chunk over $ \mathbf{x}[t - 20, t] $ and forward it through
the RNN. While this practically increases the amount of data that needs to be
processed by a factor of $ 20 $, only short sequences need to be forwarded at once and we
benefit from a high parallelization degree and comparable large minibatch size.

Additionally, one has to note that even LSTMs and GRUs can only capture a limited amount of temporal context. For instance, studies from machine translation suggest that $ 20 $ frames is a range that can be well captured by these architectures \cite{cho2014translation}. This finding is confirmed for video data in~\cite{singh16multistream}.
Also, humans usually do not need much more context to accurately classify a part of an action. Hence, storing the information of \eg frame $ 10 $ while computing the output of frame $ 500 $ is not necessary. Thus, it can be appropriate to limit the temporal scope in favor of a faster, more feasible training.

\subsection{Inference}
\label{sec:inference}

\begin{figure}[tb]
    \centering
    \includegraphics[width=0.45\textwidth]{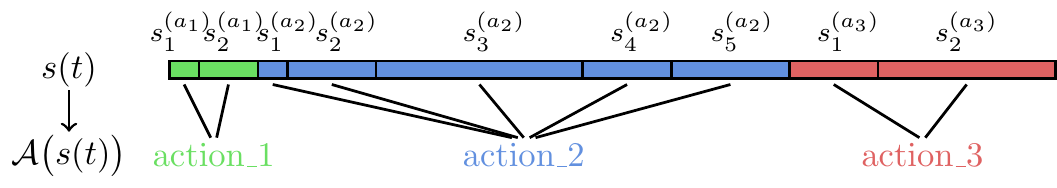}
    \caption{The extractor function $ \mathcal{A} $ computes the unique action sequence
             induced by the frame-to-subaction alignment $ s(t) $.}
    \label{fig:extractor}
    \vspace{-3mm}
\end{figure}

Based on the observation probabilities of the fine-grained subaction model and the coarse model for  overall actions, we will now discuss the combined inference of both models on video level. 

Given a video $ \mathbf{x}_1^T $, the most likely action sequence
\begin{align}
    \mathbf{\hat a}_1^N = \argmax_{\mathbf{a}_1^N} \{ p(\mathbf{x}_1^T | \mathbf{a}_1^N) \cdot p(\mathbf{a}_1^N) \}
    \label{optSequence}
\end{align}
and the corresponding frame alignment is to be found. In order to limit the amount of
action sequences to optimize over, a context-free grammar $ \mathcal{G} $ is created
from the training set as in \cite{kuehne2016weakly}.
We set $ p(\mathbf{a}_1^N) = 1 $ if $ \mathbf{a}_1^N $ is generated by $ \mathcal{G} $
and $ p(\mathbf{a}_1^N) = 0 $ otherwise. Thus, in Equation~\eqref{optSequence}, the $ \argmax $ only needs to
be taken over action sequences generated by $ \mathcal{G} $ and the factor $ p(\mathbf{a}_1^N) $ can be omitted.
Instead of finding the optimal action
sequence directly, the inference can equivalently be performed over all possible
frame-to-subaction alignments $ s(t) $ that are consistent with $ \mathcal{G} $.
Consistent means that the unique action sequence defined by $ s(t) $ is generated
by $ \mathcal{G} $. Formally, we define an extractor function $ \mathcal{A}: s(t) \mapsto \mathbf{a}_1^N $
that maps the frame-to-subaction alignment $ s(t) $ to its action sequence, see Figure~\ref{fig:extractor}
for an illustration. Equation~\eqref{optSequence} can then be rewritten as
\begin{align}
    \mathbf{\hat a}_1^N = \argmax_{s(t): \mathcal{A}(s(t)) \in \mathcal{L}(\mathcal{G})}
                          \Big\{ \prod_{t=1}^T p\big(x_t|s(t)\big) \cdot p\big(s(t)|s(t-1)\big) \Big\},
                          \label{optAlignment}
\end{align}
where $ \mathcal{L}(\mathcal{G}) $ is the set of all possible action sequences that can be generated
by $ \mathcal{G} $. Note that Equation~\eqref{optAlignment} can be solved efficiently using a Viterbi algorithm
if the grammar is context-free, see \eg \cite{jurafsky1995using}.

For training, as well as for the task of aligning videos to a given ordered action sequence
$ \mathbf{a}_1^N $, the best frame alignment to a single sequence needs to be inferred.
By defining a grammar that generates only the given action sequence
$ \mathbf{a}_1^N $, this alignment task can be solved using Equation~\eqref{optAlignment}.
For the task of temporal action segmentation, \ie when no action sequence is provided for inference,
the context-free grammar can be derived from the ordered action sequences given in the training samples.

\subsection{Training}

\begin{figure*}[tb]
    \includegraphics[width=\textwidth]{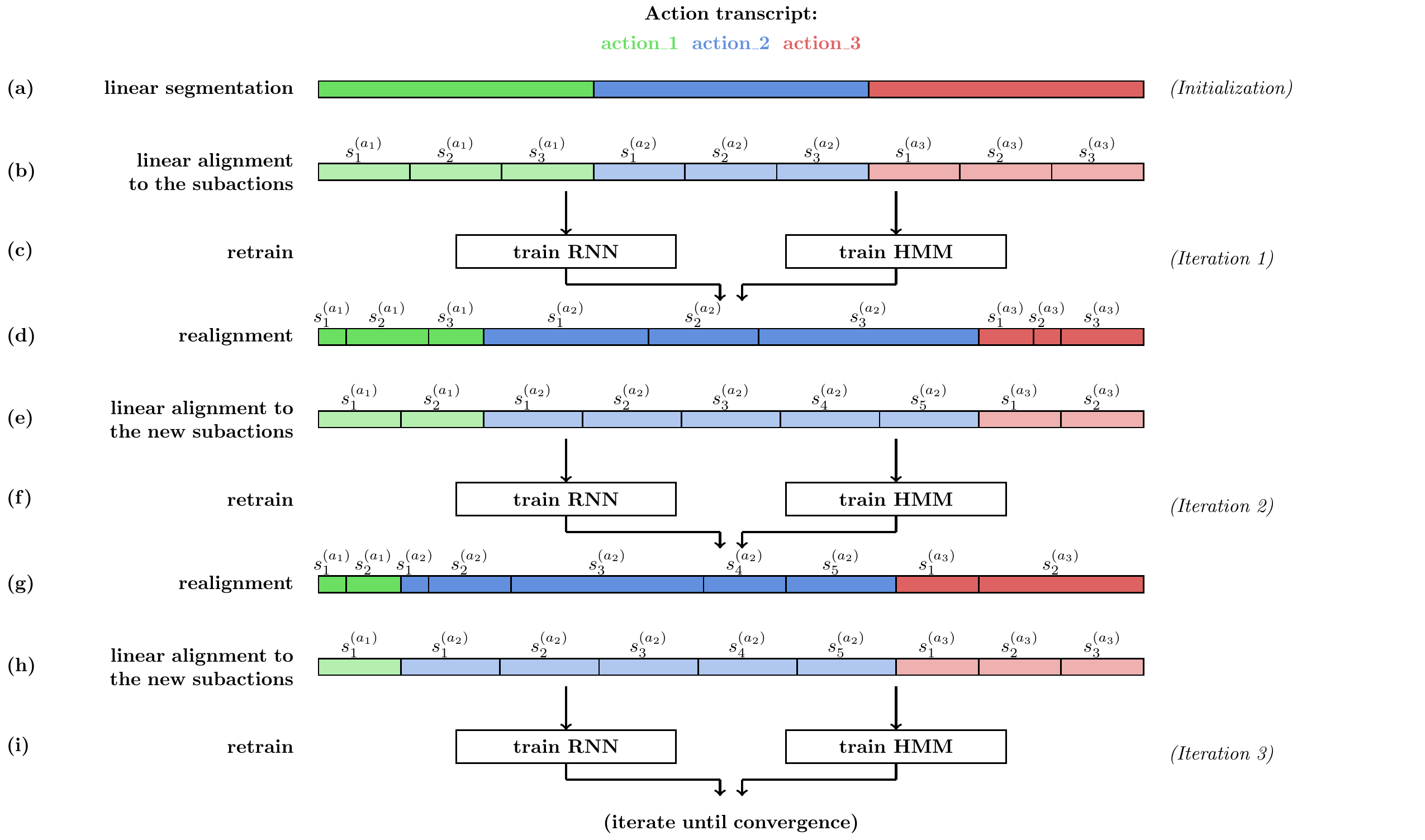}
    \caption{Training process of our model.
             Initially, each action is modeled with the same number of subactions
             and the video is linearly aligned to these subactions. Based on
             this alignment, the RNN is trained and used in combination
             with the HMMs to realign the video frames to the subactions.
             Eventually, the number of subactions per action is reestimated and the process
             is iterated until convergence.}
    \label{fig:train}
\end{figure*}

Training of the model is an iterative process, altering between both,
the recurrent neural network and the HMM training, and the alignment of frames to subaction units via
the HMM. The whole process is illustrated in Figure~\ref{fig:train}.

\noindent
\textbf{Initialization.}
The video is divided into $ N $ segments of equal size, where $ N $
is the number of action instances in the transcript (Figure~\ref{fig:train}a). 
Each action segment is further subdivided equally across the subactions (Figure~\ref{fig:train}b).
Note that this defines the mapping $ s(t) $ from frames to subactions. Each subaction should cover $ m $ frames of an action on average.
Thus, the initial number of subactions for each action is
\begin{align}
    \frac{\text{number of frames}}{\text{number of action instances} \cdot m}, \label{subactions}
\end{align}
where we usually choose $ m = 10 $ as proposed in~\cite{kuehne16end, kuehne2016weakly}.
Hence, initially each action is modeled with the same number of subactions. This can change during
the iterative optimization.


\noindent
\textbf{Iterative Training.}
The fine-grained RNN is trained with the current mapping $ s(t) $ as ground truth.
Then, the RNN and HMM are applied to the training videos and a new alignment of frames to subactions (Figure~\ref{fig:train}d) is inferred given the new fine-grained probabilities $ p(x_t|s) $ from the RNN. The new alignment is
obtained by finding the subaction mapping $ s(t) $ that best explains the data:
\begin{align}
    \hat s(t) &= \argmax_{s(t)} \Big\{ p(\mathbf{x}_1^T|\mathbf{a}_1^N) \Big\} \\
              &= \argmax_{s(t)} \Big\{ \prod_{t=1}^T p\big(x_t|s(t)\big) \cdot p\big(s(t)|s(t-1)\big) \Big\}.
              \label{realign}
\end{align}
Note that Equation~\eqref{realign} can be efficiently computed using a Viterbi
algorithm.
Once the realignment is computed for all training videos, the average length of
each action is reestimated as
\begin{align}
    \mathrm{len}(a) = \frac{\text{number of frames aligned to } a}
                           {\text{number of }a\text{-instances}} 
\end{align}
and the number of subactions is reestimated based on the updated average action lengths.
Particularly, for action $ a $, there are now $ \mathrm{len}(a) / m $ subactions, which are
again uniformly distributed among the frames assigned to the corresponding action, \cf Figure~\ref{fig:train}e.
These steps are iterated until convergence.

\noindent
\textbf{Stop Criterion.}
\label{sec:stop_criterion}
As the system iteratively approximates the optimal action segmentation on the training data, we define a stop criterion based on the overall amount of action boundaries shifted from one iteration to the succeeding one.
In iteration $ i $, let $ \mathrm{change}(i) $ denote the percentage of frames that is labeled differently compared to
iteration $ i - 1 $. We stop the optimization if the frame change rate between two iterations is less than a threshold,
\begin{align}
    |\mathrm{change}(i) - \mathrm{change}(i-1)| < \vartheta \quad \Rightarrow \quad \text{stop}.
\end{align}


\section{Experiments}

In this section, we provide a detailed analysis of our method. Code and models
are available online\footnote{https://github.com/alexanderrichard/weakly-sup-action-learning}.

\subsection{Setup}

\noindent
\textbf{Datasets.}
We evaluate the proposed approach on two heterogeneous datasets.
The Breakfast dataset is a large scale dataset with $1,712$ clips and an overall duration of $66.7$ hours. The dataset comprises various kitchen tasks such as making tea but also complex activities such as the preparation of fried egg or pancake. It features 48 action classes with a mean of $4.9$ instances per video. We follow the evaluation protocol as proposed by the authors in \cite{kuehne14language}.
 
The Hollywood extended \cite{bojanowski14weakly} dataset is an extension of the well known Hollywood dataset, featuring $937$ clips from different Hollywood movies. The clips are annotated with two or more action labels resulting in 16 different action classes overall and a mean of $2.5$ action instances per clip.

\noindent
\textbf{Features.}
For both datasets we follow the feature computation as described in \cite{kuehne16end} using improved dense trajectories (IDT) and Fisher vectors (FVs). To compute the FV representation, we first reduce the dimensionality of the IDT features from 426 to 64 by PCA and sample $150,000$ randomly selected features to build a GMM with 64 Gaussians. The Fisher vector representation \cite{sanchez13image} for each frame is computed over a sliding window of 20 frames. Following \cite{perronnin10improving}, we apply power and l2 normalization to the resulting FV representation. Additionally, we reduce the final FV representation from $8,192$ to 64 dimensions via PCA to keep the overall video representation manageable and easier to process. 

\begin{figure}[t]
    \centering
    \includegraphics[scale=0.8]{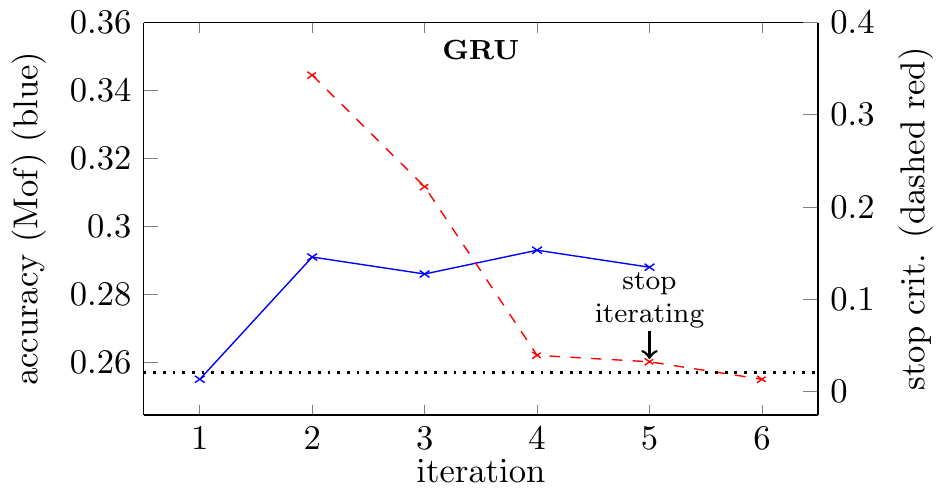}
    \\
    \includegraphics[scale=0.8]{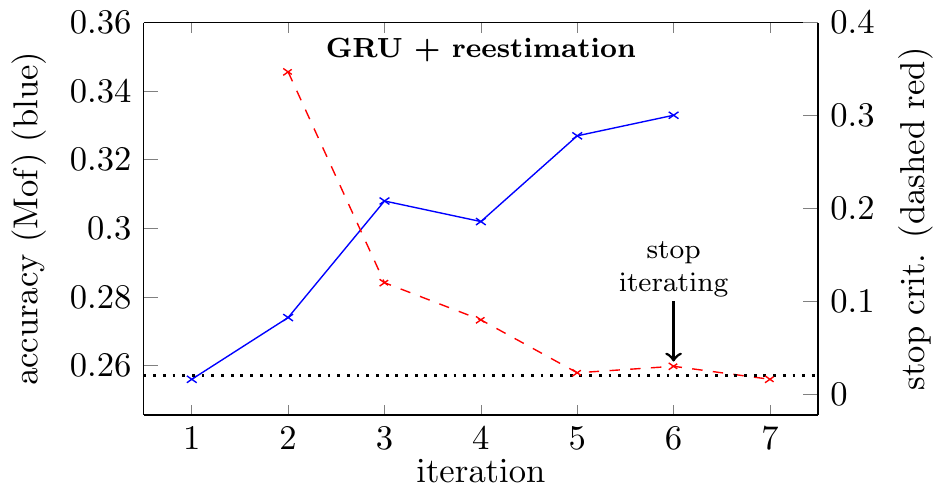}
    \caption{Stop criterion for two experiments, one using the fine-grained model and no subaction reestimation,
             one using it with subaction reestimation. The blue curve shows the frame accuracy over the iterations,
             the red curve shows the frame change rate between the current and the preceding iteration. The dashed line
             represents the threshold $ \vartheta = 0.02 $.}
    \label{fig:stopcrit}
    \vspace{-3mm}
\end{figure}

\noindent
\textbf{Stop Criterion.}
For the stop criterion, we fix $ \vartheta = 0.02 $, \ie if the difference of the frame change between two iterations
is less than two percent, we stop iterating. Figure~\ref{fig:stopcrit} illustrates the criterion for two example experiments.
The blue curve is the frame accuracy, the red curve is the difference of frame changes between two iterations. It can be seen
that after a few iterations, the frame accuracy does not increase anymore but begins to oscillate, see also
Table~\ref{tab:low_level_eval}. Comparing the frame change rate of the train data is a good indicator of when to stop iterating.
In all experiments, we compute results based on the alignment of the last iteration before the threshold $ \vartheta $ is crossed.

\subsection{Analysis of the Coarse Model}
\label{sec:analysis_coarse}

 \begin{table}[t] \footnotesize
    \centering
   \begin{tabularx}{0.45\textwidth}{Xr}
        \toprule
        Breakfast                            & Accuracy (Mof)  \\
        \midrule           
        \textit{GRU no subactions}        &  $  22.4 $  \\
        \textit{GRU w/o reestimation}    &  $  28.8 $  \\
        \textit{GRU + reestimation}      &  $  33.3 $  \\
        \midrule
       \textit{GRU + GT length}           &  $  51.3 $  \\
        \bottomrule
    \end{tabularx}
    \caption{Results for temporal segmentation on the Breakfast dataset comparing accuracy of the proposed system (GRU + reestimation) to the accuracy of the same architecture without subactions (GRU no subactions) and to the architecture with subclasses but without reestimation. } 
    \label{tab:reest_eval}
    \vspace{-3mm}
\end{table}

\begin{figure*}[t]
    \centering
    \includegraphics[width=0.48\textwidth]{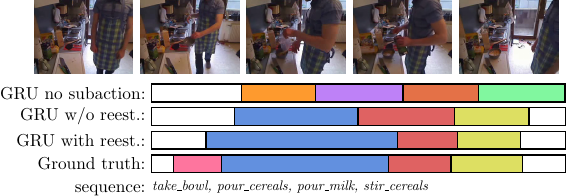}
    \hfill
    \vspace{3mm}
    \includegraphics[width=0.48\textwidth]{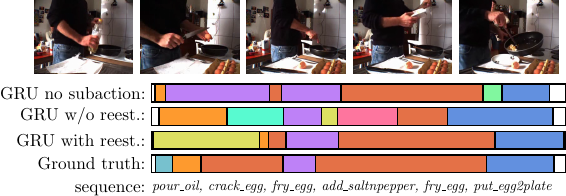}
    \caption{Example segmentation results for two samples from the Breakfast dataset showing the segmentation result for ´``preparing cereals'' and ``preparing friedegg''. Although the actions are not always correctly detected, there is still reasonable alignment of detected actions and ground truth boundaries.}
    \label{fig:exp_seg}
    \vspace{-3mm}
\end{figure*}

For the following evaluation of our system, we report performance for the task of temporal action segmentation, \ie the combined video segmentation and classification. Given a video without any further information, the task is to classify all frames according to their related action. This includes to infer which actions occur in the video, in which order they occur, and their respective start and end frames. We evaluate on the test set of the Breakfast dataset and report results as mean accuracy over frames (Mof) (see \cite{kuehne14language}). We iterate the system until the stop criterion as described in Section~\ref{sec:stop_criterion} is reached. 

First, we regard the properties of the coarse action modeling. We therefore compare the proposed system to the results of the same setting, but without further subdividing actions into subactions (GRU no subactions, Table~\ref{tab:reest_eval}). Additionally, we regard results of the system without reestimation of subactions during optimization (GRU w/o reestimation, Table~\ref{tab:reest_eval}). For the system without reestimation, we follow the initial steps as shown in Figure \ref{fig:train}, thus, we linearly segment the videos according to the number of actions, generate an initial subaction alignment, train the respective subaction classes, and realign the sequence based on the RNN output. But, opposed to the setup with reestimation, we omit the step of reestimating the number of subclasses and the following alignment. Instead we just use the output of the realignment to retrain the classifier and iterate the process of training, alignment, and re-training. Thus, the number of subclasses is constant and the coarse model is not adapted to the overall estimated length of the action class.
Finally, we compare against a system in which we use the ground truth boundaries to compute the mean length of an action class and set the number of subactions based on the mean ground truth length (GRU + GT length, Table~\ref{tab:reest_eval}). Here, all action classes are still uniformly initialized, but longer action classes are divided into more subactions than shorter ones. We include this artificial scenario as it models the performance in case that the optimal number of subaction classes would be found.

One can see in Table \ref{tab:reest_eval} that recognition performance without subactions is significantly below all other configurations, supporting the idea that subaction modeling in general helps recognition in this scenario. 
The model with subactions, but without reestimation, improves over the single class model, but is still below the system with subaction reestimation. Compared to that, the model with subaction reestimation performs 5\% better. 
We ascribe the performance increase of the reestimated model to the fact that a good performance is highly related to the correct number of subactions, thus to a good length representation of the single actions. By reestimating the overall number of subactions after each iteration, this ground truth length is approximated. The impact of the number of subactions becomes clear, when considering the results when the ground truth action lengths are used. Here the performance of the same system, just with different numbers of subactions, increases by almost $20\%$. 
Qualitative results on two example videos are shown in Figure~\ref{fig:exp_seg}.

\subsection{Analysis of the Fine-Grained Model}

 \begin{table}[t] \footnotesize
    \centering
   \begin{tabularx}{0.45\textwidth}{Xrrrrr}
        \toprule
        Breakfast & Iter 1 & Iter 2 & Iter 3  & Iter 4  &  Iter 5   \\
        \cmidrule(lr){1-6}           
        \textit{GMM w/o reest.}  &  $  15.3 $ & $  23.3 $ & $  26.3 $ & $  27.0 $ & $  26.5 $  \\
        \textit{MLP w/o reest.}  &  $  22.4 $ & $  24.0 $ & $  23.7 $ & $  23.1 $ & $  20.3 $  \\
        \textit{GRU w/o reest.}  &  $  25.5 $ & $  29.1 $ & $  28.6 $ & $  29.3 $ & $  28.8 $  \\
        \cmidrule(lr){1-6}
        \textit{GRU w/o HMM}     &  $  21.3 $ & $  20.1 $ & $  23.8 $ & $  21.8 $ & $  22.4 $  \\
        \bottomrule
    \end{tabularx}
    \caption{Results for low level recognition with an MLP compared to GRUs over five iterations.
             The MLP quickly starts to overfit, whereas the GRU oscillates at a constant level.
             Last row: A GRU without HMM, showing that short term dependencies are well captured
             by the fine-grained recurrent network.}
    \label{tab:low_level_eval}
    \vspace{-3mm}
\end{table}

In order to analyze the capability of capturing temporal context with the recurrent network, we compare it to a system
where a Gaussian mixture model (GMM) and a multilayer perceptron (MLP) are used instead. Both only operate on frame level and do
not capture fine-grained information between the frames. In order to provide a fair comparison to the recurrent model,
we equip the MLP with a single hidden layer of rectified units  such that it has the same number of parameters as the recurrent
network.
We use a simplified version of the system without subaction reestimation to achieve comparable results after each iteration.

Results for the first five iterations on the Breakfast dataset are shown in Table \ref{tab:low_level_eval}.
We find that GRUs clearly outperform both, GMMs and MLPs, starting with $25.5\%$ for the initial recognition, and reaching up to $29.3\%$ after the fourth iteration. Particularly, the MLP baseline stays continuously below this performance. Thus, it can be assumed that the the additional information gained by recurrent connections in this context supports classification.
One can further see that the MLP reaches its best performance after the second iteration and then continuously decreases, whereas the GRU begins to oscillate around $29\%$, hinting that the MLP also starts to overfit at an earlier stage compared to the GRU.
In the last row of Table~\ref{tab:low_level_eval}, the coarse model, \ie the HMM, is removed from the system.
Thus, there is no modeling of subactions anymore and the GRU directly learns the original action classes.
The performance is clearly worse than with the coarse model, but still the system is remarkably good, being on par with the
MLP that uses the coarse model (second row of Table~\ref{tab:low_level_eval}).

\subsection{Comparison to State-of-the-Art}

\textbf{Temporal Action Segmentation.}
 \begin{table}[t] \footnotesize
    \centering
   \begin{tabularx}{0.45\textwidth}{Xrr}
        \toprule
               & \textbf{Breakfast} & \textbf{Hollywood Ext.} \\
        \midrule
         Model &  Accuracy (Mof)    & Jacc. (IoU) \\
        \midrule           
        OCDC \cite{bojanowski14weakly}*     &  $ 8.9 $  & - \\
        HTK \cite{kuehne2016weakly}         &  $ 25.9 $ & $ 8.6 $ \\
        ECTC \cite{huang2016connectionist}  &  $ 27.7 $ & - \\
        \midrule 
         GRU w/o reestimation               &  $ 28.8 $ & $ 11.2 $ \\
         GRU + reestimation                 &  $ \mathbf{33.3} $ & $ \mathbf{11.9} $  \\
       \bottomrule
    \end{tabularx}
    \caption{Comparison of temporal action segmentation performance for GRU based weak learning with other approaches. For the Breakfast dataset, we report performance as mean over frames (Mof), for Hollywood extended, we measure the Jaccard index as intersection over union for this task (*from \cite{huang2016connectionist}).} 
    \label{tab:sota1}
    \vspace{-3mm}
\end{table}
We compare our system to three different approaches published for the task: The first is the Ordered Constrained Discriminative Clustering (OCDC) proposed by Bojanowski \etal \cite{bojanowski14weakly}, which has been introduced on the Hollywood extended dataset. Second, we compare against the HTK system used by Kuehne \etal \cite{kuehne2016weakly}, and third against the Extended Connectionist Temporal Classification (ECTC) by Huang \etal \cite{huang2016connectionist}. 
For the Breakfast dataset, we follow the evaluation script of \cite{kuehne14language, huang2016connectionist} and report results as mean accuracy over frames over four splits. 
For the Hollywood Extended dataset, we follow the evaluation script of \cite{kuehne2016weakly} and report the Jaccard index (Jacc.) as intersection over union (IoU) over 10 splits. Results are shown in Table~\ref{tab:sota1}.

One can see that the proposed subaction based GRU systems show a good performance, and that both subaction based systems outperform current approaches on the two evaluated datasets.
It also shows that the GRU based system without reestimation shows comparable performance to other RNN based systems, such as ECTC by \cite{huang2016connectionist}, which uses an LSTM model with comparable size. The significant increase in accuracy of our method can be attributed to the reestimation.
We also observe that the performance boost of the system with reestimation is more prominent on the Breakfast than on the Hollywood extended dataset. We attribute this to the fact that in case of Hollywood extended, all action classes usually have a consistent mean frame length, whereas in case of Breakfast, the mean length of action classes significantly varies. Thus, the benefit of adapting to action class lengths increases with the heterogeneity of the target action classes.

\textbf{Action Alignment.}
 \begin{table}[t] \footnotesize
  \centering
   \begin{tabularx}{0.45\textwidth}{Xrr}
        \toprule
               & \textbf{Breakfast} & \textbf{Hollywood Ext.} \\
        \midrule
         Model &  Jacc. (IoD)       & Jacc. (IoD) \\
        \midrule           
        OCDC \cite{bojanowski14weakly}  &  $ 23.4 $  & $ 43.9 $ \\
        HTK \cite{kuehne2016weakly}**   &  $ 40.6 $  & $ 42.4 $ \\
        ECTC \cite{huang2016connectionist}** & -     & $ 41.0 $ \\ 
        \midrule 
         GRU w/o reestimation           &  $ 41.5 $  & $ 45.6 $ \\
         GRU + reestimation             &  $ \mathbf{47.3} $ & $ \mathbf{46.3} $  \\
       \bottomrule
    \end{tabularx}
    \caption{Results for action alignment on the test set of the Breakfast and the Hollywood extended dataset reported as Jaccard index of intersection over detection (IoD) (**results obtained from the authors).} 
    \label{tab:sota2}
    \vspace{-3mm}
\end{table}
We also address the task of action alignment. Here, we assume that given a video and a sequence of temporally ordered actions, the task is to infer the respective boundaries for the given action order. We report results for the test set of Breakfast as well as for the Hollywood extended dataset based on the Jaccard index (Jacc.) computed as intersection over detection (IoD) as proposed by \cite{bojanowski14weakly}. The results are shown in Table~\ref{tab:sota2}.\footnote{Errata: The results for Hollywood Extended have been corrected. $ 45.6 $ and $ 46.3 $ are the correct numbers, the results $ 50.1 $ and $ 51.1 $ from the original CVPR paper are incorrect.}
Here, the GRU system without reestimation performs on par with other systems for the alignment task on the Breakfast dataset, but the GRU system with reestimation again shows a clear improvement over current systems.


\section{Conclusion}

We presented an approach for weakly supervised learning of human actions based on a combination of a discriminative representation of subactions, modeled by a recurrent neural network, and a coarse probabilistic model to allow for a temporal alignment and inference over long sequences. Although the system itself already shows good results, the performance is significantly improved by approximating the number of subactions for the different action classes. Accordingly, we propose to adapt the number of subaction classes by iterating realignment and reestimation during training. The resulting models outperform state-of-the-art on various weak learning tasks such as temporal action segmentation and action alignment.

\vspace{1mm}
\noindent
\textbf{Acknowledgments.}
The work has been financially supported by the DFG projects KU 3396/2-1 (Hierarchical Models for Action Recognition and Analysis in Video Data) and
GA 1927/4-1 (DFG Research Unit FOR 2535 Anticipating Human Behavior) and the ERC Starting Grant ARCA (677650).


{\small
\bibliographystyle{ieee}
\bibliography{references}
}

\end{document}